\def\year{2021}\relax
\documentclass[letterpaper]{article} 
\usepackage{aaai21}  
\usepackage{times}  
\usepackage{helvet} 
\usepackage{courier}  
\usepackage[hyphens]{url}  
\usepackage{graphicx} 
\usepackage{natbib}  
\usepackage{caption} 

\usepackage[switch]{lineno}
\usepackage{microtype}
\usepackage{subfigure}
\usepackage{subfig}
\usepackage{booktabs} 
\usepackage{amsmath}
\usepackage{cancel}
\usepackage{xcolor}
\usepackage{algorithm}
\usepackage{algorithmic}
\usepackage{eqparbox}
\usepackage{soul}

\newcommand\blfootnote[1]{%
  \begingroup
  \renewcommand\thefootnote{}\footnote{#1}%
  \addtocounter{footnote}{-1}%
  \endgroup
}

\nocopyright

\title{A Bayesian Approach to Identifying Representational Errors}
\author {
Ramya Ramakrishnan\textsuperscript{\rm 1}$^*$,
Vaibhav Unhelkar\textsuperscript{\rm 2}$^*$,
Ece Kamar\textsuperscript{\rm 3},
Julie Shah\textsuperscript{\rm 4}\\
}

\affiliations {
    \textsuperscript{\rm 1} ASAPP \\
    \textsuperscript{\rm 2} Rice University \\
    \textsuperscript{\rm 3} Microsoft Research \\
    \textsuperscript{\rm 4} Massachusetts Institute of Technology \\
    rramakrishnan@asapp.com, vaibhav.unhelkar@rice.edu, eckamar@microsoft.com, julie\_a\_shah@csail.mit.edu
}

\begin{document}
\maketitle

\begin{abstract}
Trained AI systems and expert decision makers can make errors that are often difficult to identify and understand. Determining the root cause for these errors can improve future decisions. This work presents Generative Error Model (GEM), a generative model for inferring representational errors based on observations of an actor's behavior (either simulated agent, robot, or human). The model considers two sources of error: those that occur due to representational limitations -- ``blind spots'' -- and non-representational errors, such as those caused by noise in execution or systematic errors present in the actor's policy. Disambiguating these two error types allows for targeted refinement of the actor's policy (i.e., representational errors require perceptual augmentation, while other errors can be reduced through methods such as improved training or attention support). We present a Bayesian inference algorithm for GEM and evaluate its utility in recovering representational errors on multiple domains. Results show that our approach can recover blind spots of both reinforcement learning agents as well as human users.\blfootnote{$^*$Both authors contributed equally. Work was done while both authors were affiliated with MIT.}
\end{abstract}

\section{Introduction}
\label{intro}
Errors in complex decision tasks can be frequent, and the cause is often unclear, making it difficult to take informed steps towards reducing them. For example, a self-driving car may make mistakes on the road due to several factors. One factor might be the limited sensing capability of the car, which hinders its ability to view and act appropriately. Given accurate perception, other factors could also cause errors, such as a lack of training or limited memory. 

In this work, we present Generative Error Model (GEM), a generative model that identifies task errors that occur due to representational deficiencies. We consider a setting in which a flawed actor provides demonstrations of a task. This actor can be a trained RL agent, a robot, a human, or any decision-making system. In fact, our experiments consider one domain with an RL agent and another with human users, which highlights the versatility of our approach. This actor may make many different types of errors and an observer's goal is to determine the errors that are due to limitations in the actor's perception of the world. To do this, our approach separates these representational errors from errors caused by imperfections in the actor's policy. Learning this distinction is useful because if an error is due to an actor's flawed representation of the world (e.g., an RL agent's representation of the state may be missing features), we can augment the actor's perception with technology such as an indicator that gives a notification when there is a car outside of the driver's view, enabling them to ``perceive'' what they otherwise could not. If an error is not due to a representational limitation (i.e., the state of the world has an accurate representation), an assistive agent can help reduce other errors through, for example, customized training, reminders, or attention support \cite{horvitz2013methods}.

\begin{figure}
    \centering
    \includegraphics[width=0.35\textwidth]{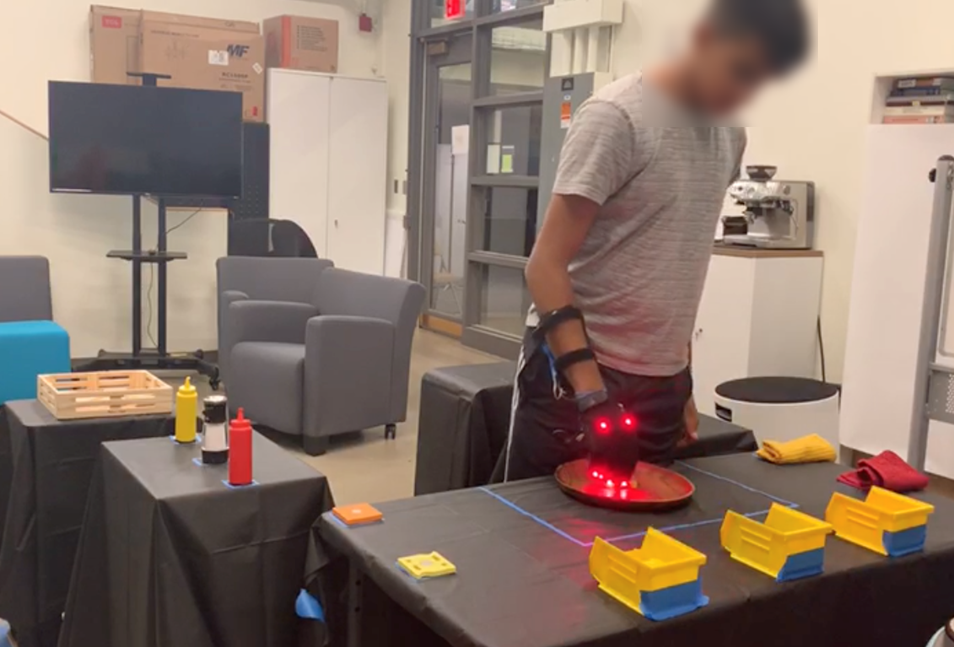}
    \caption{A participant prepares dishes in our kitchen task. Our generative model is used to determine representational human errors in this setting.}
    \label{fig:kitchen_motivation}
\end{figure}

In our problem setup, the observer that is reasoning about actor errors has full knowledge of the given world, including the true representation and optimal policy. The observer has access to demonstration data from the actor performing the task, as well as labels indicating when errors occurred. To identify representational errors, we propose a generative model GEM that encodes two latent factors: ``blind spots'' -- attributes of the true representation that are unobservable by the actor -- and ``execution noise -- errors that may occur despite a correct representation. We chose these two latent sources for error because if representation is flawed, all decision-making based on it will also be flawed. Identifying the actor's representation first allows the observer to then better reason about decisions/errors of the actor given that representation. GEM works by estimating the actor's potentially flawed view of the world and the expected optimal action given that view. One benefit of this model is that it can be personalized to different types of actors, as each can have different observational deficiencies and noise levels.

To infer representational errors using GEM, we perform Bayesian inference using exact techniques as well as approximate methods (collapsed Gibbs sampling). Results on two domains (a gridworld task with a trained RL agent and a kitchen task with real human users) indicate that the model can successfully infer representation limitations (i.e., which features are not observable by the actor) by separating them from execution noise. This generative framework is flexible, and can be augmented based on domain-specific assumptions about observable variables or conditional independencies. The GEM graphical model allows us to better capture the inner decision processes that might affect actions and consequently, errors. Iteratively identifying and reducing these errors through such approaches can be extremely beneficial for improving safety in the real world.

\section{Related Work}

\noindent \textbf{Reinforcement learning:} Model-based and model-free reinforcement learning (RL) can be used to learn effective policies when performing a task \cite{abbeel2004apprenticeship,ziebart2008maximum,merel2017learning,stadie2017third,hadfield2016cooperative}. When trying to explain agent errors, it is helpful to understand agent behaviors and mistakes, which involve modeling causal knowledge about the world and the learned policy. Both model-free and model-based RL involve modeling causal knowledge \cite{gershman2017reinforcement} through hidden state inference. Partially Observable Markov Decision Processes \cite{kaelbling1998planning,thrun2000monte,hoelscher2018utilizing,ross2008online,silver2010monte,ng2000pegasus} (POMDPs) are also relevant, as they assume agents cannot directly observe the true state and instead maintain a distribution over possible beliefs. These methods, however, are used to generate behavior, rather than understand behavior. Some work \cite{sunberg2017value,sadigh2016information,javdani2018shared} has used POMDPs to infer agent intents, but they do not focus on identifying representational blind spots.

Prior works in multi-agent RL \cite{littman1994markov,tan1993multi,gupta2017cooperative,iqbal2019coordinated,raileanu2018modeling,torrey2013teaching} have developed methods for working with several agents whose beliefs and observations are unknown; however, the majority of these works have focused on generating behavior to improve joint team performance rather than understanding the behavior and errors of other agents. Approaches used to understand other's behaviors exist, including methods in learning from demonstration and inverse RL \cite{argall2009survey,arora2018survey,zhifei2012survey,ng2000algorithms,abbeel2004apprenticeship}, but these do not consider errors and often assume positive examples.

\noindent \textbf{Human cognition:} Prior cognitive science literature \cite{van2009understanding,baker2014modeling,pantelis2014inferring,saxe2005against} has explored models that simulate decision-making processes of humans. Works in this space can be useful for identifying the reasons behind human decisions and errors. Griffiths et al \cite{griffiths2010probabilistic} argued that top-down probabilistic models can better generalize and represent human decisions compared with bottom-up connectionist approaches. In another work \cite{griffiths2006optimal}, the authors used errors in human predictions to better understand people's assumptions about the world and their decision-making processes. 

Several authors have also developed methods to better understand how agents model other agents. Ullman et al \cite{ullman2009help} proposed a method to infer an agent's intention, focused on whether an agent is helping or hindering the achievement of another agent's goals. Baker et al \cite{baker2006bayesian,baker2009action} developed a Bayesian framework to explain how people predict another agent's actions based on their observations of task execution. Overall, understanding human cognition is quite relevant to our work at a high level, but our ultimate goal is different, as we would like to specifically identify representational errors. Our method is also flexible enough to work for humans as well as machines/agents.\\

\noindent \textbf{Representation learning:} Learning appropriate state representations and identifying when a representation is insufficient \cite{bengio2013representation,radford2015unsupervised,diuk2008object,wang2015deep} is also relevant to this work, because we consider a setting in which an actor and an observer operate with different representations, and the observer must identify errors occurring from the actor's flawed representation. Many works \cite{devin2018deep,silva2018object,li2018object,ramakrishnan2016interpretable,diuk2008object} have used object-based representations as a natural way to describe the world, resulting in improved transfer and interpretability. Many RL works \cite{jaderberg2016reinforcement,ma2018universal,lyu2019sdrl,nachum2018near,haarnoja2018latent} have introduced approaches to automatically learn representations that generalize well across tasks. Overall, representation learning is focused on learning a low-dimensional representation and not necessarily errors.

Some works have specifically addressed flawed representations; for example, in one work \cite{unhelkar2018learning}, Bayesian nonparametric techniques were used to learn unmodeled features of an agent with limited representation. Identifying agent blind spots \cite{ramakrishnan2018discovering,ramakrishnan2019overcoming} similarly assumes that a flawed agent learns from an oracle demonstrator. These assume the reverse scenario: that an agent learns about its own representational deficiencies using expert feedback, while we consider the scenario of identifying representation limitations of another agent.

\section{Generative Error Model (GEM)}
\label{sec:error_model}

\begin{figure}
\begin{center}
\includegraphics[width=0.45\textwidth]{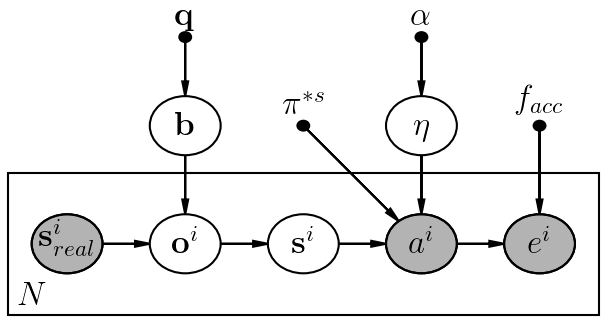}
\end{center}
\caption{Generative Error Model (GEM): A graphical model of the generative process for errors.}
\label{fig:models}
\end{figure}

Learning the latent causes of errors is essential for reducing and resolving them. Thus, we provide a generative model of decision-making, which enables us to encode the causal process due to which errors were generated. We focus on two main error sources: representational and execution errors. Representational errors (or ``blind spots'') result from an inability to observe critical attributes of the given task; execution errors occur when an actor has the correct representation of the world but still makes mistakes due to other factors, such as limited practice, slow reaction times, carelessness, and random noise during execution.

In our problem setup, an observer observes an actor perform a task and aims to determine the source of the actor's errors. The observer is assumed to know the true representation of the world, which includes all task attributes that are needed to optimally act in the given task, and the optimal policy itself. This knowledge in turn allows the observer to reason about the actor's errors.

The input to our model is a set of $N$ demonstrations from the actor: $D = \{(\textbf{s}^i_{real},a^i,e^i)\}, i \in [1,...,N]$, represented as a list of state-action-error tuples. These demonstrations can be from a trained RL policy, a human actor, a robot controller, etc. The true state of the world, $\textbf{s}^i_{real} = [f_1, f_2 ..., f_k]$, is represented as a vector of features. The actor takes an action, $a^i \in \mathcal{A}$, and an error indicator, $e^i \in \{0,1\}$, provides information about when errors occurred. We assume that the observer has access to an acceptable function $f_{acc}(s,a)$, which returns 1 if action $a$ is acceptable in state $s$, and 0 otherwise. This is a more relaxed function than just checking for optimality and can be defined by the observer flexibly based on the domain. For example, $f_{acc}$ can be based on how much worse the actor's decision is from the optimal. 

Given these demonstrations, we want to identify errors caused by representational limitations. We assume that the actor may not be able to see the true state of the world $\textbf{s}^i_{real}$. Instead, the actor receives an observation $\textbf{o}^i$, which is a (potentially many-to-one) function of $\textbf{s}^i_{real}$. We define the transformation function, which maps the true state to the observation, using a random variable denoting the actor's blind spots. In general, this transformation can be an arbitrary nonlinear, stochastic function. In this formulation, we represent the blind spot as a vector $\textbf{b} = [b_1, ..., b_k], \textbf{b}_j \in \{0,1\}$, which encodes information about the features the actor cannot observe (i.e., when $\textbf{b}_j = 1$, the actor cannot observe feature $f_j$ of the true state). Each value $\textbf{b}_j$ is sampled from a Bernoulli distribution with hyperparameter $q$. Given true state $\textbf{s}^i_{real}$ and blind spot vector $\textbf{b}$, we can obtain the actor's possibly flawed observation, $\textbf{o}^i$. The generative process is as follows:\vspace{-0.2cm}

\begin{flalign*}
\textbf{b}_j \sim \text{Bernoulli}(\textbf{q})\\
\textbf{o}^i \sim Pr(\textbf{o}^i | \textbf{s}^{i}_{real}, \textbf{b}) \text{    }  \forall i
\end{flalign*}

We model the blind spot $\textbf{b}$ as a mask over $\textbf{s}^i_{real}$ in order to compute $\textbf{o}^i$. Thus, in our case, this transition is deterministic; however, generally, the observation can be represented as an arbitrary, blind spot-dependent transformation of the state.

Now that we have the actor's observation of the world, we want to explain the actor's decisions in order to understand when these decisions cause errors. The observer has access to the optimal policy $\pi^{*s}: \mathcal{S}_{real} \rightarrow \mathcal{A}$ with respect to the true state of the world $\textbf{s}_{real}\in \mathcal{S}_{real}$. However, since the actor's observation and the true state do not share the same representation, the observer cannot directly query the optimal policy with the observation. To bridge this gap, we propose estimating the actor's implicit view of the world in terms of the true state representation, where values for missing features denote the actor's assumptions about those features. This variable captures this assumption of the world, which will ultimately affect the actor's action selection. Our model also allows us to capture the case where the actor truly does not know and thus does not make any implicit assumption. 

The model maps the actor's observation $\textbf{o}^i$ to a distribution over possible implicit states $\textbf{s}^i$. Explicitly modelling the actor's view of the world helps to reason about actor decisions. Given $\textbf{s}^i$ and our known optimal policy $\pi^{*s}$, the observer can compute the optimal actor decision. However, an actor may not always act optimally, so we include an additional variable $\eta$ that denotes noise in execution. Noise can be represented using an arbitrary distribution. In this work, we model $\eta$ with discrete values, ranging from 1\% to 40\%, in 5\% intervals (e.g., expert: 1\% noise, beginner: 40\% noise). This variable intuitively captures how often the actor deviates from the optimal policy given $\textbf{s}^i$. Thus, the action is derived by sampling from the optimal policy with probability $1-\eta$ and a random action with probability $\eta$. The error is computed using the known acceptable function. We sample actions and errors as follows. Figure \ref{fig:models} depicts the full graphical model.\vspace{-0.4cm}

\begin{flalign*}
\textbf{s}^i \sim Pr(\textbf{s}^i | \textbf{o}^i) \text{    }  \forall i\\
\eta \sim \text{Multinomial}(\alpha)\\
a^i \sim \pi^{*s}(a^i | \textbf{s}^i, \eta) \text{    } \forall i\\
e^i \sim f_{acc}(e^i | \textbf{s}^i_{real}, a^i) \text{    } \forall i
\end{flalign*}
The observer's goal is to learn $P(\textbf{b}, \eta | D)$: the probability of both the blind spot vector $\textbf{b}$ and the noise parameter $\eta$ given demonstration data $D$. The output is a probability distribution over the blind spot vectors and execution noise, which can be used to analyze the actor's errors.

\section{GEM Inference}

We use techniques from Bayesian inference to infer latent variables $\textbf{b}$ and $\eta$ given demonstration data $D$: $P(\textbf{b}, \eta | D)$. In domains with small state spaces, where exact inference is possible, variable elimination is used to eliminate each variable. Based on this approach, the following equations, included below, are used to infer the two latent variables. The sum is computed directly using the final equation to obtain the posterior distribution:\vspace{-0.2cm}

\begin{align*}
\label{eq:variable_eliminate}
& P(\textbf{b}, \eta | D) = \frac{P(D | \textbf{b}, \eta) P(\textbf{b}, \eta)}{P(D)}\\
& = \frac{P(\textbf{b})P(\eta) \prod\limits_i \sum\limits_{\textbf{o}^i} P(\textbf{o}^i | \textbf{s}^i_{real}, \textbf{b}) \sum\limits_{\textbf{s}^i}P(\textbf{s}^i | \textbf{o}^i)\pi^{*s}(a^i | \textbf{s}^i, \eta)}{\sum\limits_{\textbf{b},\eta} P(\textbf{b})P(\eta) \prod\limits_i \sum\limits_{\textbf{o}^i} P(\textbf{o}^i | \textbf{s}^i_{real}, \textbf{b}) \sum\limits_{\textbf{s}^i}P(\textbf{s}^i | \textbf{o}^i)\pi^{*s}(a^i | \textbf{s}^i, \eta)}
\end{align*}

For more complex domains, approximate inference is used. In collapsed Gibbs sampling, as shown in Algorithm \ref{fig:gibbs}, we alternate between sampling the blind spot vector and the noise vector given all other variables. The variables $\textbf{o}^i$ and $\textbf{s}^i$ are collapsed because they are highly correlated with $\textbf{b}$. Sampling one given the other variables results in very little variation among the sampled values. Thus, $\textbf{o}^i$ and $\textbf{s}^i$ can be marginalized out so that only $\textbf{b}$ is sampled. Given the value of $\textbf{b}$, $\eta$ can be sampled, again marginalizing out the other two variables. After this sampling process, the probability of $\textbf{b}$ and $\eta$ can be computed based on the sampled data, which gives an estimate of the posterior distribution.

\begin{minipage}{0.46\textwidth}
\centering
\begin{algorithm}[H]
\centering
\caption{Collapsed Gibbs Sampling}
\label{fig:gibbs}
\begin{algorithmic}
\STATE $\textbf{Given  }$ data $D = \{\textbf{s}^i_{real}, a^i, e^i\}$, samples $S = []$
\FOR{$(\textbf{s}^i_{real}, a^i, e^i) \in D$}
\STATE $\textbf{b} \sim P(\textbf{b})$ \COMMENT{Initialize blind spot vector}
\STATE $\eta \sim P(\eta)$ \COMMENT{Initialize noise parameter}
\FOR{$num \in [1,...,N]$}
\STATE $\textbf{b} \sim P(\textbf{b} | D, \eta)$ \COMMENT{Sample blind spot given rest}
\STATE $\eta \sim P(\eta | D, \textbf{b})$ \COMMENT{Sample noise given rest}
\STATE $S += (\textbf{b}, \eta)$ \COMMENT{Include sample}
\ENDFOR
\ENDFOR
\STATE $P(\textbf{b}, \eta) = \frac{n_{(\textbf{b}, \eta)}}{|D|}$ \COMMENT{Compute probabilities}
\end{algorithmic}
\end{algorithm}
\end{minipage}\\\\

\section{Experiments}
\label{sec:experiments}

\begin{figure*}
\begin{center}
\begin{minipage}{0.25\textwidth}
\includegraphics[width=\textwidth]{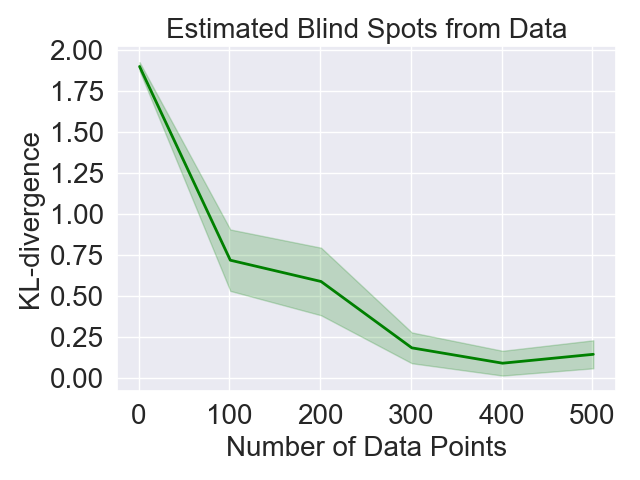}
\end{minipage}\hfill
\begin{minipage}{0.25\textwidth}
\includegraphics[width=\textwidth]{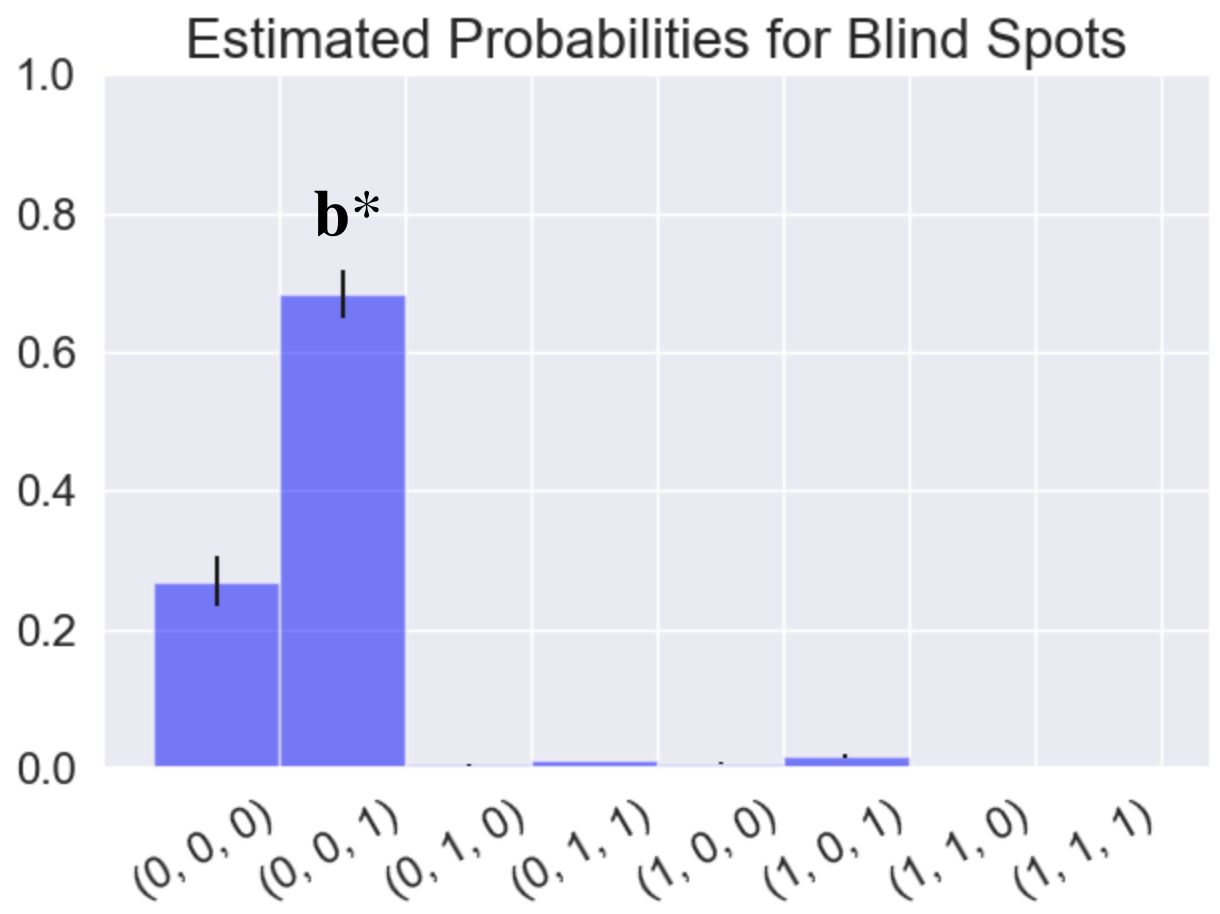}
\end{minipage}\hfill
\begin{minipage}{0.25\textwidth}
\includegraphics[width=\textwidth]{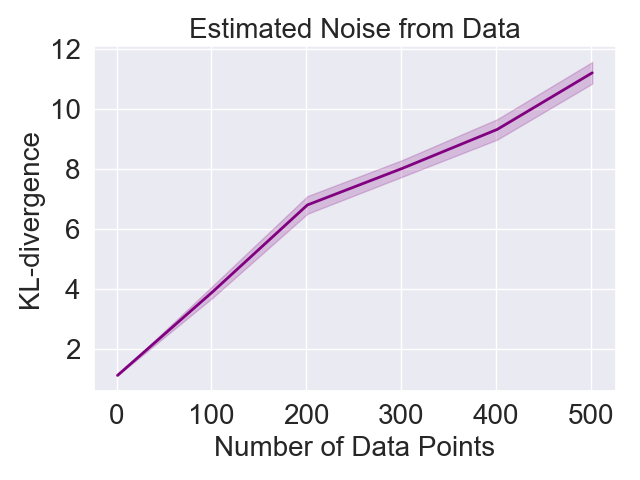}
\end{minipage}\hfill
\begin{minipage}{0.25\textwidth}
\includegraphics[width=\textwidth]{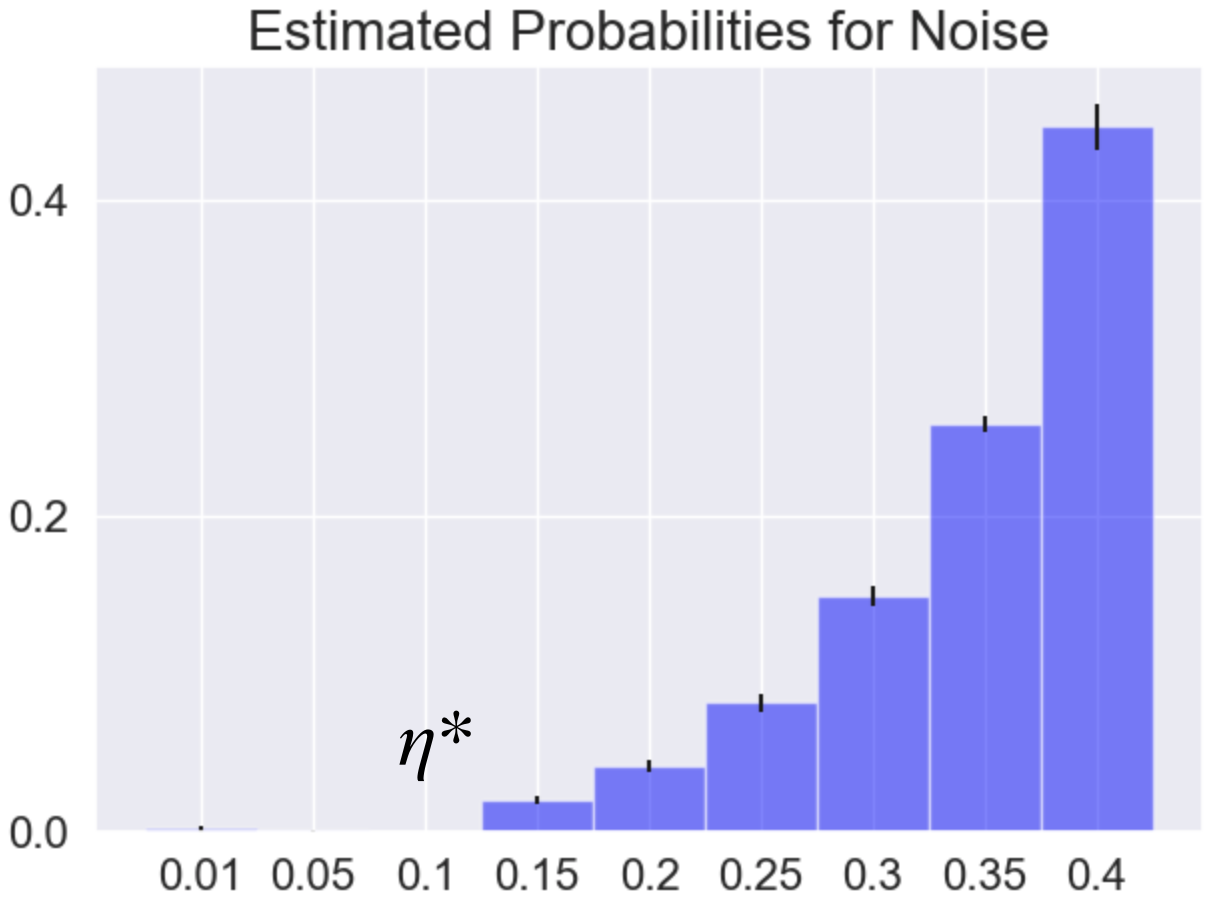}
\end{minipage}\hfill
\end{center}
\caption{Estimated blind spots given demonstration data. Vector (0,0,1) means the object color feature is a blind spot for the actor.}
\label{fig:budget_results}
\end{figure*}

We evaluated our approach on two domains. The first is a gridworld environment, where we trained an RL agent that could not distinguish between differently colored objects. Our approach enabled inference of these blind spots. The second domain is a kitchen task, in which real human users were instructed to study a menu of several dishes and prepare these dishes from memory. Salt and sugar were purposefully made indistinguishable from one another, which resulted in consistent systematic errors. Further, due to the difficulty of the task, people made other mistakes, such as forgetting to include some of the ingredients in a given dish. Our findings indicate that our approach can infer representational human errors in this domain, which can inform the redesign of the kitchen environment to provide a more accurate view of the world. We discuss details of our experiments below.

\subsection{Gridworld domain}

We first conducted experiments in a gridworld domain: specifically, a 10x10 grid with one object, colored either green or red, placed randomly in one of the 100 locations. An agent was tasked with collecting green objects and avoiding red objects. The state was represented as: $[\Delta x, \Delta y, c]$, denoting the x-distance from the object, the y-distance from the object, and the object color respectively. We simulated a color-blind agent that could not tell the difference between green and red. We included noise in the policy by randomly sampling actions with $10\%$ probability.

\subsubsection{Our approach inferred blind spots}

Our model inferred the blind spot vector given data from demonstrations. We generated data from an RL agent who could not observe color and implicitly assumed that any given object was green. We ran variable elimination to estimate $P(\textbf{b},\eta|D)$, averaged over 100 runs. The first plot in Figure \ref{fig:budget_results} plots the variation of the KL-divergence between the estimated $\textbf{b}$ and the true distribution with respect to the size of the demonstration data. The estimate of $\textbf{b}$ approached the true vector as the amount of data was increased, as shown by the decreasing KL-divergence. The histogram depicts the estimated probabilities for each blind spot vector when the data consisted of 100 datapoints. The model correctly predicted the true vector (0,0,1) with 70\% probability; in other words, the model was able to recognize that the agent likely observed the x- and y-distances away from the object but not the object's color.

\subsubsection{Noise was more challenging to estimate}

Our model had a more difficult time estimating the noise parameter compared to inferring the blind spots. In this domain, we included a range of execution noise from 1\% noise to 40\% with intervals of 5\%. The third plot in Figure \ref{fig:budget_results} depicts the KL-divergence of the estimated $\eta$ and the true distribution. The true agent acted with 10\% noise, but our approach inferred 40\% noise as the most likely because a higher (but incorrect) value of execution noise better explains the errors in the data. In other words, 40\% error allows for more randomness in action decisions than 10\% does so the model can more easily generate the demonstration data. This suggests the need for regularization and more informative priors for noise estimation, potentially based on domain expertise. This finding raises the question of whether adding the noise parameter is useful if the model cannot estimate it precisely.

\subsubsection{Modeling noise improved blind spot inference}

\begin{figure}
\begin{center}
\includegraphics[width=0.4\textwidth]{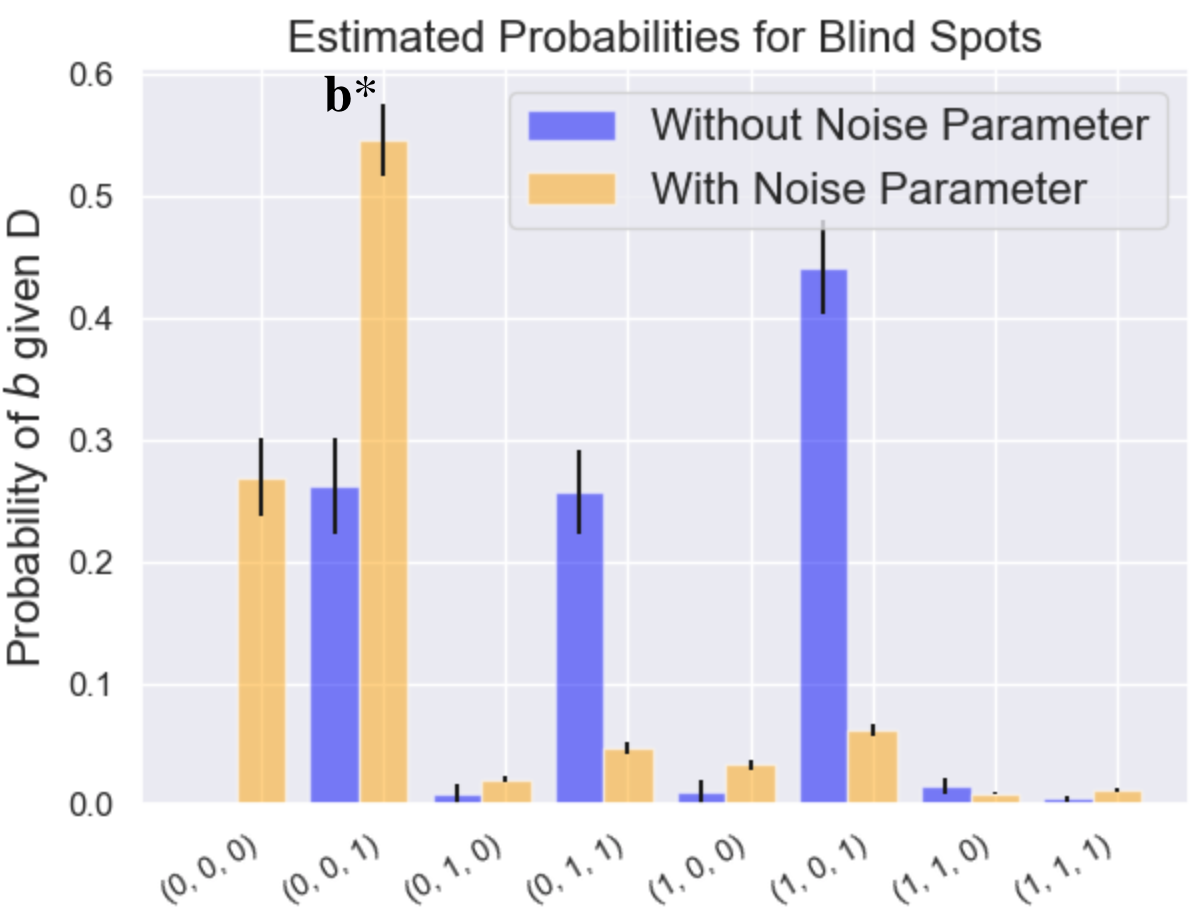}
\end{center}
\caption{Adding the noise parameter resulted in more accurate estimates of blind spots in the presence of noisy demonstration data.}
\label{fig:noise}
\end{figure}

We next compared a version of our model without an explicit noise parameter (we included a fixed $\eta$=0.01 deviation to avoid probabilities of 0) to our original model that estimates $\eta$, averaged over 100 runs. The fixed-noise baseline is included to analyze the benefit of including noise in the model and inferring it. Figure \ref{fig:noise} shows that when the agent acted with a high amount of noise (30\%), the model without $\eta$ attributed the noisy actions to confused estimates for the blind spots. For example, if the agent's actions did not match the optimal for a large percentage of states, this led the observer to think that the agent had additional blind spots. 

The model that inferred $\eta$ was able to explain much of this noise and recover the true blind spot vector. This shows that while it is difficult to exactly estimate this parameter, inferring $\eta$ provides greater flexibility for the model, resulting in better blind spot estimates. In order to improve the estimate of $\eta$, more informative priors for execution noise can be incorporated.

\subsubsection{Estimated implicit state representation captured agent assumptions about missing features}

Another interesting insight from our model is the most likely estimate of $\textbf{s}^i$ for a given data point, which provides more information than just recovering the actor's blind spots because it estimates the actor's assumption about missing observational features. For example, blind spots might indicate that the actor does not observe color, while the implicit state $\textbf{s}^i$ tells us the actor is likely assuming the color to be green.

\begin{figure}
\begin{center}
\includegraphics[width=0.38\textwidth]{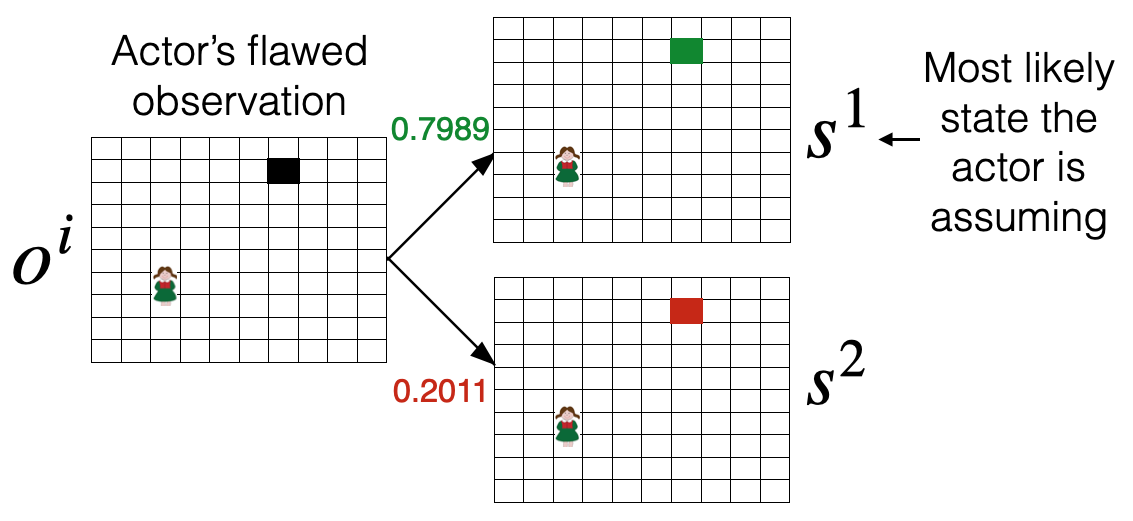}
\end{center}
\caption{The most likely implicit state the agent may be assuming in order to make action decisions.}
\label{fig:s_h}
\end{figure}

The model can be used to compute $P(\textbf{s}^i | D)$, which is a distribution over possible $\textbf{s}^i$ values given the data. The highest probability $\textbf{s}^i$ captures information about which assumptions the agent might be making about the features it does not observe. As Figure \ref{fig:s_h} indicates, our model estimated that the agent's actions matched with about 80\% probability to an optimal policy assuming the object is green. Note that this could either mean that the agent assigned a probability every time and made a decision under uncertainty or that the agent assumed green eight out of 10 times and red the remaining two. Our model does not distinguish between the two cases. In this way, our generative model allows for a more in-depth analysis of an actor's decision-making process.

\subsection{Kitchen domain}

The second evaluation was on a kitchen domain. For this application, real human users were tasked with preparing dishes in a mock kitchen environment. They were given two minutes to study a menu of 3 dishes and then made 25 dishes from memory based on randomly generated orders from the menu, with a 1-minute refresher of the menu every 5 dishes. The task of memorizing and quickly making dishes was quite challenging, so we expected users to make many mistakes, some because they did not know the identity of certain ingredients (blind spots) and others because of limited memory or carelessness (execution noise). To create a representational blind spot in the task, we intentionally included salt and sugar in recipes without disambiguating them in the kitchen. From this data collection, we obtained a total of 2,420 state-action-error tuples. The proposed model was then used to identify the cause of human errors. 

\subsubsection{Data included representation and memory errors}

The state for this task was constructed with the following features: 1 feature representing the type of dish the participant was currently making out of the possible three, 7 binary features representing whether each of the ingredients necessary for the given dish had been included, and 14 features denoting the ingredient's position at each of the 14 locations throughout the kitchen. The action at each time step was to either select one of the 14 ingredients or to ``serve'' the dish once the person thought it was complete. If the selected ingredient was part of the specified dish and had not yet been used, then the participant's action was defined as acceptable. The participant was considered to have made an error if she included an ingredient not meant for the given dish, or served a dish with missing ingredients.

We modelled the human's blind spot vector as a binary vector with the same size as the state, denoting whether the human could observe each state feature. Because the space of blind spot vectors was huge, we restricted the human's possible blind spots to include only the features denoting which ingredient was present at each kitchen location and allowed up to three ingredients to be unobservable to the person. The implicit state $\textbf{s}^i$ represented the human's estimate of the true world state and was mainly used to predict which ingredients the human assumed were present at each location. Because the state space was larger for this domain, we used collapsed Gibbs sampling for inference (Algorithm \ref{fig:gibbs}).

To evaluate our model, we computed an estimate of the ground truth based on our task setup. There were two ground-truth blind spots, the location of the salt and the location of the sugar, both of which were unobservable to the person. We obtained the noise ground-truth value by removing the salt/sugar errors and taking the percentage of the remaining erroneous actions. While these were the aggregate ground truths, each participant had varying levels of noise during task execution, making it difficult to truly recover this blind spot vector. We observed a total error rate of approximately $24\%$ in the demonstration data, with $7.23\%$ attributed to blind spot (salt/sugar) mistakes and $16.78\%$ resulting from other forms of noise. With so much noise in the data, identifying the true blind spot vector was challenging.

\begin{figure}
\begin{center}
\includegraphics[width=0.38\textwidth]{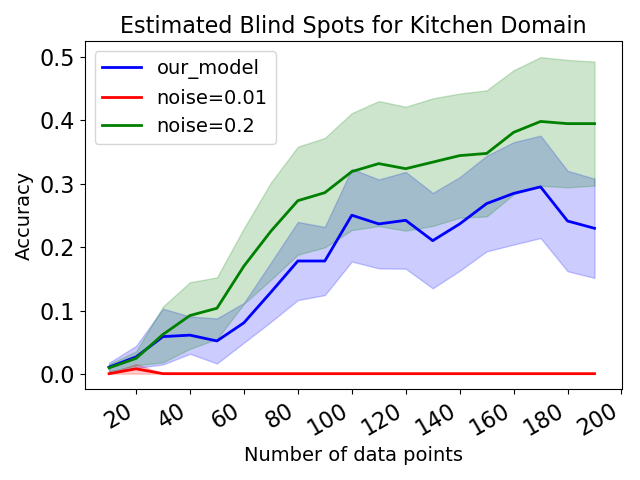}
\end{center}
\caption{Performance of our approach on predicting human blind spots in the kitchen domain.}
\label{fig:kitchen_bs}
\end{figure}

\subsubsection{Our approach correctly identified expected blind spots in the kitchen task}

In Figure \ref{fig:kitchen_bs}, we plot our model's performance when predicting blind spots as the demonstration data budget increased. We ran the model for each participant individually, and the input was the first $n$ datapoints in a given participant's data (e.g., 60 datapoints along the x-axis indicate that the model used the first 60 $(\textbf{s}^i_{real}, a^i, e^i)$ tuples for one participant in order to infer blind spots).

We measured the accuracy of predicting the true blind spot vector -- an extremely challenging task. Selecting a vector randomly at chance would result in an accuracy of 0.002 due to the size of the space of possible vectors. Our model was able to achieve approximately 30\% accuracy of predicting the ground truth blind spot vector $\textbf{b}$, which is quite difficult. Recovering the most likely blind spot vector requires separating the noise in the real data to determine consistent mistakes a participant might make due to a partial view of the world.

In order to further analyze the benefits of our model, we compared our model, which infers both the blind spot and noise parameters to a baseline model that only infers blind spots with a fixed, constant noise value. In one condition, we set the noise to a minimal value of $\eta=0.01$ ($1\%$ noise) -- similar to that present in the gridworld experiments. This scenario resulted in very poor performance because the model consistently attributed noisy human actions to additional blind spots since it lacked any other way to explain the randomness in human actions. We also included a baseline with a fixed $20\%$ noise level, which is similar to an oracle that knows approximately the true value of noise. Our GEM model was able to automatically infer the two ground-truth blind spots with higher accuracy than the fixed $\eta=0.01$ baseline and achieved performance close to the oracle variant. In real-world applications, it is unlikely for the true amount of execution noise to be known, so it is best to infer both variables automatically, as our model does. This also allows the model to learn personalized noise values for each user.

\begin{figure}
\begin{center}
\includegraphics[width=0.35\textwidth]{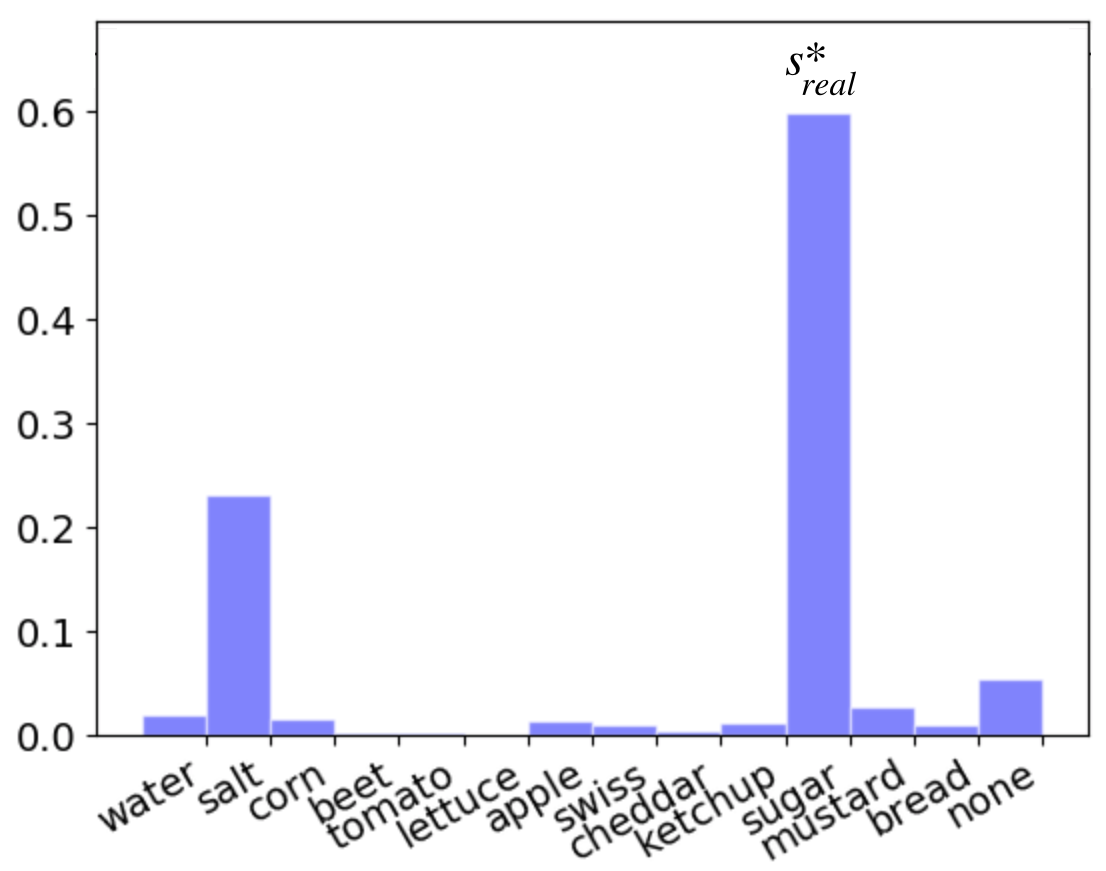}
\end{center}
\caption{Estimate of participants' implicit state of the world on the kitchen domain.}
\label{fig:kitchen_s_h}
\end{figure}

\subsubsection{Our model was able to capture participants' implicit assumptions}

Since our model includes latent variables intended to explain human actions, we can also query for other quantities, such as the most likely human's implicit state of the world $\textbf{s}^i$. Similar to the gridworld task, this variable can be inferred by calculating $P(\textbf{s}^i|D)$. Then, marginalization can be used to determine the probability of various values for a specific feature. Figure \ref{fig:kitchen_s_h} plots a distribution over possible ingredients that the human assumed were located in the true salt location in the kitchen. The model predicted sugar as the most likely ingredient, matching our intuitive expectation. Thus, our model can serve as a tool to better understand human errors and the assumptions leading to those errors.

\section{Discussion}

These findings indicate that our approach can infer blind spots on the two tested domains. Here, we discuss limitations of the approach and directions for next steps. First, this work assumes that an observer has access to the representation of the world and the optimal/acceptable policy. While this is realized in several real-world applications (e.g., an AI pilot is trained in a simulator with limited observability, while a human observes all features of the simulation and can identify weaknesses in the pilot's performance), our approach can be combined with other methods for addressing flawed representations to handle a larger range of problems \cite{ramakrishnan2018discovering,unhelkar2018learning}. 

\begin{figure}
\begin{center}
\includegraphics[width=0.4\textwidth]{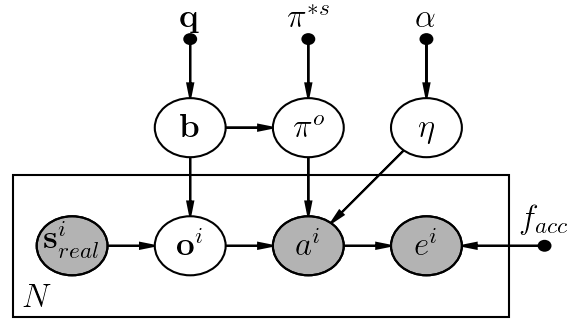}
\end{center}
\caption{A variation of the original model that directly estimates the actor's policy in terms of the actor's observation.}
\label{fig:model2}
\end{figure}

Secondly, the GEM model provides one way of inferring representational errors. However, the model forces us to estimate the actor's view of the world in terms of the observer's state representation. Explaining the actor's behavior through this intermediate representation may not be appropriate in some settings; for example, if the actor cannot perceive a given color, she may not be filling it in with some implicit color, but instead acting directly according to the flawed observation. In order to model this, we present a variation of our graphical model (Figure \ref{fig:model2}) that estimates the actor's policy directly with respect to the representation of the observation $\pi^o: \mathcal{O} \rightarrow \mathcal{A}$. The computational cost of performing inference with this model is much higher ($|\textbf{b}| + N|\textbf{o}| + N|\textbf{s}|$ parameters for the original vs. $|\textbf{b}| + N|\textbf{o}| + |\pi^o|$ parameters for the variation where $|\pi^o|$ can be large). However, it can express a richer and more complex model of decision-making and merits exploration in future work.

In both of these models, since there are many factors that can affect action decisions, disentangling the different error sources from one another can be challenging. For example, if significant noise exists within the demonstration data, it may be difficult to determine whether there is a systematic blind spot with limited noise or no blind spot with very high noise. As there are multiple possible explanations for the data, the problem can be ill-posed in certain applications. In such cases, it is important to include strong priors for different variables within the model to increase the chance of identifying the most likely explanation for the errors. It could also be useful to include additional causes for errors in the model (although this could make disentangling even more difficult). For example, the random noise factor in this model encompasses errors resulting from various factors, including carelessness, limited training, an incorrect model of the world, and others. Separating errors that occur because a person does not know the task objective from those that are truly random can be useful for enabling more informed fixing of these errors.

Next, the model depends upon the representation of each variable. For example, in the kitchen domain, our model learned that people's perceptions of the salt and sugar were incorrect. Specifically, they confused the locations because the ingredients were visually indistinguishable. To learn this, the representation of the state, blind spot, and observation had to include the locations of the ingredients. One possible extension of the model is to have the blind spot depend on the true state (e.g., for images, it may be better for the blind spot to be state-dependent, which would represent the actor's inability to see a region of that particular image). Another example is relaxing binary blind spots to be continuous, where each variable can take a value between $[0,1]$ (e.g., a dimly lit region might not entirely be visible to a person). 

Depending on the domain, our model is also flexible enough to make changes to conditional independencies and observed variables. For example, in this work, the observer had access to whether the actor's actions led to an error during the task, but perhaps the error is not directly available and instead, the observer has access to a noisy observation of the error. Another possibility is that the error is only observed at the end of a sequence of actions, which would require propagating error signals to earlier decisions. 

Understanding errors can ultimately lead to targeted refinement of an actor's representation and policy. For representational deficiencies, we can add new sensors to enable the actor to observe what they originally could not. Execution errors can be addressed through techniques, such as training procedures, attention support, and reminders. An interesting future direction is to extend this model to evaluate the benefit of different improvement strategies for assisting the actor. For example, we can select the intervention that results in the largest improvement in task performance. This iterative process of identifying and fixing errors can improve decision-making on complex tasks.

\section{Conclusion}
In this work, we present a generative model GEM for identifying errors of an actor (e.g., agent, human, robot) caused by representational limitations. Our approach models and infers the actor's estimate of the true state in order to explain her action decisions. Bayesian inference is used to separate blind spots, or representational deficiencies, from execution noise, which represents how often actions deviate from the optimal policy. We demonstrate through experiments on a gridworld domain with a trained RL agent and a kitchen task with real user data that we are able to identify blind spots. We additionally use the generative model to recover the actor's view of the world, which provides more clarity into why certain actions were taken. In future work, we plan to augment the model with additional components to better express complex decision-making processes and use the findings from this model to reduce errors and improve safety on real-world tasks.


\bibstyle{aaai21}
\bibliography{paper}

\end{document}